# Automated identification of hookahs (waterpipes) on Instagram: an application in feature extraction using Convolutional Neural Network and Support Vector Machine classification

Youshan Zhang, M.S., [1] Jon-Patrick Allem, Ph.D., M.A., [2] Jennifer B. Unger, Ph.D.[2]

Tess Boley Cruz, Ph.D.[2]

[1] Department of Computer Science, Lehigh University, Bethlehem, PA, USA,

[2] Keck School of Medicine, University of Southern California, Los Angeles, CA,

USA

**Address for correspondence:** Jon-Patrick Allem, 2001 N. Soto Street, 3rd Floor

Mail, Los Angeles, CA 90032. Phone: (858) 603-0812, E-Mail: allem@usc.edu

**Word count**: 2643 Tables: 0 Figures: 11

**Keywords:** Image classification; Feature extraction; support vector machine; convolution neural network; Instagram; social media.

**Abstract**

**Background:** Instagram, with millions of posts per day, can be used to inform public health surveillance targets and policies. However, current research relying on image-based data often relies on hand coding of images which is time consuming and costly, ultimately limiting the scope of the study. Current best practices in automated image classification (e.g., support vector machine (SVM), Backpropagation (BP) neural network, and artificial neural network) are limited in their capacity to accurately distinguish between objects within images. **Objective:** This study demonstrates how convolutional neural network (CNN) can be used to extract unique features within an image and how SVM can then be used to classify the image. **Methods:** Images of waterpipes or hookah (an emerging tobacco product possessing similar harms to that of cigarettes) were collected from Instagram and used in analyses (n=840). CNN was used to extract unique features from images identified to contain waterpipes. A SVM classifier was built to distinguish between images with and without waterpipes. Methods for image classification were then compared to show how a CNN + SVM classifier could improve accuracy. **Results:** As the number of the validated training images increased, the total number of extracted features increased. Additionally, as the number of features learned by the SVM classifier increased, the average level of accuracy increased. Overall, 99.5% of the 420 images classified were correctly identified as either hookah or non-hookah images. This level of accuracy was an improvement over earlier methods that used SVM, CNN or Bag of Features (BOF) alone. **Conclusions:** CNN extracts more features of the images allowing a SVM classifier to be better informed, resulting in higher accuracy compared with methods that extract fewer features. Future research can use this method to grow the scope of image-based studies. The methods presented here may be able to help detect increases in popularity of certain tobacco products over time on social media. By taking images of waterpipes from Instagram, we place our methods in a context that can be utilized to inform health researchers analyzing social media to understand user experience with emerging tobacco products and inform public health surveillance targets and policies.

# 1 Introduction

Instagram, with millions of posts per day, [1] can be used to inform public health surveillance targets and policies. However, current research relying on image-based data often relies on hand coding of images, [2][3] which ultimately limits the scope of the study. Images from social media may be more useful than findings from text-based platforms alone (e.g., Twitter, Reddit) when attempting to understand health behaviors e.g., user experiences with emerging tobacco products.[4] While automated image classification is useful for large-scale image classification (e.g., processing and assigning labels to millions of images), current best practices in automated image classification are limited in their capacity to accurately distinguish between objects within images[5][6][7]. Automated image classification has been used in supervised, unsupervised, and hybrid approaches in classifying data[8][9][10]. Compared with unsupervised methods, supervised methods can be divided into stages of training and testing. The training stage consists of training a classifier by images and its labels e.g., describing image content, such as person, dog, elephant, etc. The testing stage predicts the labels of the test images (in a new set of images) by the trained classifier.

Prior research has focused on ways to overcome the methodological challenges of automated image classification such as low accuracy. For example, Pettronnin and colleagues improved the Fisher Kernel approach to extend the bag-of-visual-words (also called bag-of-features (BOF)) for large-scale image classification using internet images from ImageNet and Flickr, which increased precision from 47.9% to 58.3%, but did not improve accuracy[5]. Kesari and colleagues used the Backpropagation neural network approach to classify large images with good accuracy (97.02%), but this approach could not identify multiple categories of an image[6]. To reduce the time and spatial complexity of images, Simonyan and colleagues proposed two visualization techniques using deep Convolutional Networks (ConvNets) to classify artificial images. They combined understandable visualizations of ConvNets, maximizing the scores of images within different classes with gradient-based ConvNets visualization generating the saliency map (also called features map, which can represent the influence of pixels in image on image classification results) of every image (corresponding to one class) in order to use a deconvolution (also called transpose of convolution, which performs upsampling tasks instead of downsample in convolution layer ) network to segment objects in the images[7].

These earlier approaches have moved automated image classification forward; however there are still a number of significant limitations to overcome [11][12][13]. For example, the large number of images that need to be extracted to train a model requires great computational power. Additionally, the BOF method cannot localize the objects within an image and cannot use visual word positions (e.g., if a cup was in an image, BOF could not find its position)[14][15]. SVMs have a limitation in showing the transparency of results since the final model is difficult to visualize. It is also a challenge to choose a suitable kernel in kernel SVM [16][17][18]. CNN, on the other hand, can improve the generalization of the algorithm and can solve nonlinear problems.  While CNN has high accuracy, to get better results, the parameters should be fine-tuned (e.g., input image size, patch size, and the number of convolution layers), and network performance is hard to optimize [19][20].

The purpose of this study is to determine whether combining CNN and SVM can achieve higher accuracy in image classification compared to CNN or SVM alone. To this end, data from Instagram containing images of waterpipes also known as hookah (an emerging tobacco product possessing similar harms to that of cigarettes) were examined. By taking data from Instagram, we place our methods in a context that can be utilized to later inform researchers in the health domain who wish to analyze social media to understand user experience with emerging tobacco products and inform public health surveillance targets and public policies [2][3][4][21][22][23][24][25].

## 2 Methods

### 2.1 Data acquisition

Data used in this study comprised posts on Instagram between February 19, 2016 and May 19, 2016 in the United States that included the hashtag #hookah. A total of n = 820 images was used in this study. The ground truth was manually labeled (hookah and non-hookah images).  To balance the data and classes, the training images include 420 images (210 hookah and 210 non-hookah images), test images also included 420 images (210 hookah and 210 non-hookah images). Further details on data collection are described elsewhere[24]. Matlab was used to classify images into two categories (images containing a waterpipe (hookah) and those without).

### 2.2 Convolution neural network

Image features comprised of 25 layers were extracted using AlexNet [26][27][28](a well-trained convolution neural network software). The architecture of AlexNet is shown in Figure 1. Among these 25 layers, there are input and output layers, seven Rectified Linear Units (ReLU) layers, two normalization layers, three pooling layers, two dropout layers (drop), one softmax layer (prob) and eight learnable weights layers which contain five convolutional layers (conv) and three fully connected layers (fc) (refer to [26] for more details). The input layer is comprised of 227×227-pixel images. The ReLU layer reduces the number of epochs to achieve the training error rate greater than traditional tanh units. The normalization layer (norm) increases generalization and reduces the error rate. The pooling layers summarize the outputs of adjacent pooling units [29]. The dropout layer efficiently decreases the test errors,[30] and both dropout layer and the softmax layer reduce the over-fitting phenomenon, while the output layer is the categories of images. To extract the features, we fine-tuned the network by removing the last two layers of the original 25 layers, since all of the layers are not suitable for extracting the features. The layers at the beginning of the network can only detect the edges of the images, so we used the results of the fully connected layers to extract features.

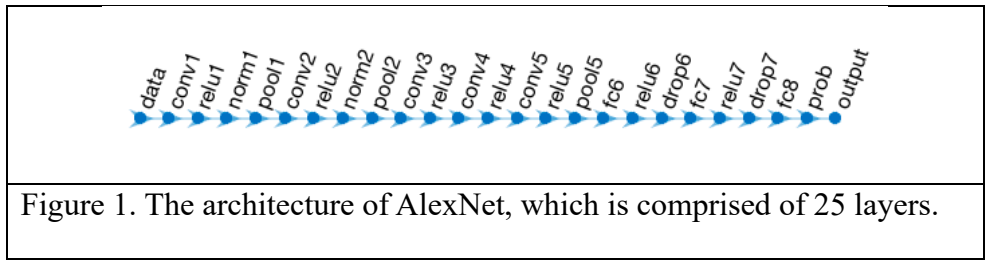

Figure 1. The architecture of AlexNet, which is comprised of 25 layers.

**2.3 Support Vector Machine**

SVM, a supervised learning model with algorithms that analyze data for classification, has been used to predict the categories of objects in images[31][32]. Our proposed method goes beyond earlier research as the input (feature vectors) was based on the results of the convolutional neural network, which can boost accuracy and save time. AlexNet was used to extract features and those features were used to then train the SVM classifier, requiring only minutes to train all images, therefore, saving time (see [11] for more details about SVM). Once the SVM classifier was trained using the feature vectors, the categories of images were predicted.

**2.4 Analytical approach**

First, we classified images into two categories: hookah and non-hookah images and labeled accordingly. Figure 2 shows the classification scheme e.g., the input image dimension is 227×227×3 pixels and the output of CNN is the 4096×1×1 feature maps, which are used to train the SVM classifier, then the classifier is used to predict the categories (hookah vs. non-hookah) of test images. The hookah images contain a waterpipe and the non-hookah images do not contain a waterpipe (Figure 3). Next, we divided image sets into training and test images; the training images were used to extract and learn the features (N=420, randomly selected), while the test images were used to calculate the accuracy of the method (N=420, randomly selected). To extract features of the images, the dimension of the input images was made uniform e.g., the image size was 227×227, since the image dimensions of 227x227 are default of AlexNet. If the image is larger or smaller, we resized the dimensions of the input image to 227x227. We loaded the pre-trained CNN by utilizing AlexNet[26], which has been trained by more than one million images. As discussed above, AlexNet was fine-tuned in our method e.g., we removed the last two layers of the AlexNet and used the data of the final fully connected layer. Based on the data of the last fully connected layer, we computed the features of the training and test images based on CNN. Then the class labels were extracted from the training and test image sets.

To optimize the SVM classifier, we automatically optimized hyperparameters (such as learning rate, the number of layers in CNN, mini-batch size) of the waterpipe features vector, and based on the optimized results, we arrived at an optimized SVM classifier (Figure 2)[32][33][34]. We then assessed the performance of the SVM classifier by using the test images and increased the number of images to improve accuracy (the number of images increased from 42 to 420) (Figures 3 & 4). Based on the trained classifier, we predicted the classes of new images.

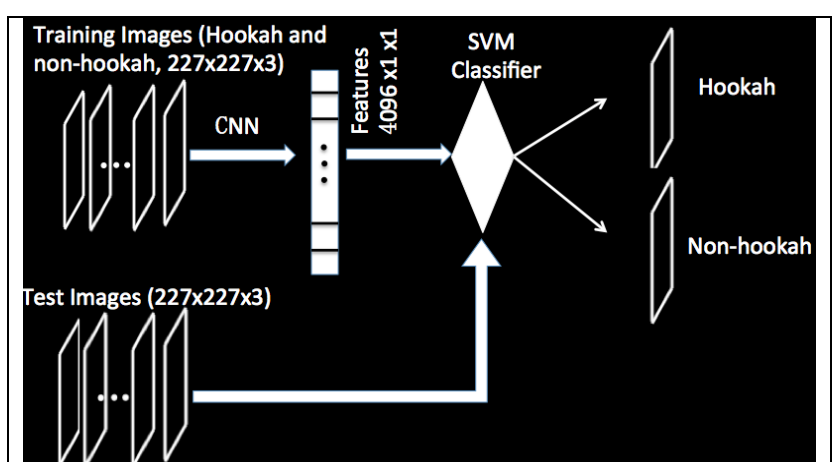

Figure 2 shows the scheme of our method. Input images dimension is 227×227×3. The output of CNN was 4096×1×1 features maps of two image classes. These features were trained by the SVM classifier, and the trained classifier was later used to predict the categories of test images.

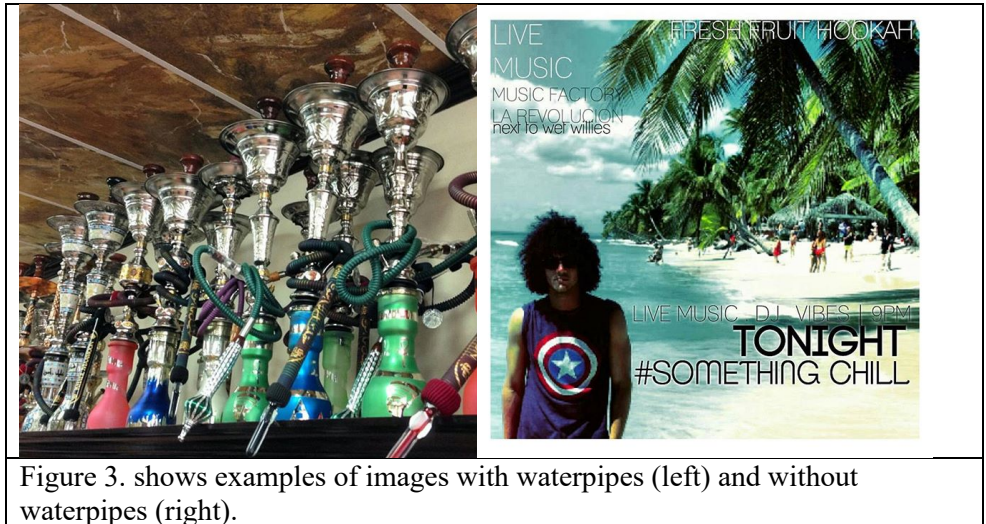

Figure 3. shows examples of images with waterpipes (left) and without waterpipes (right).

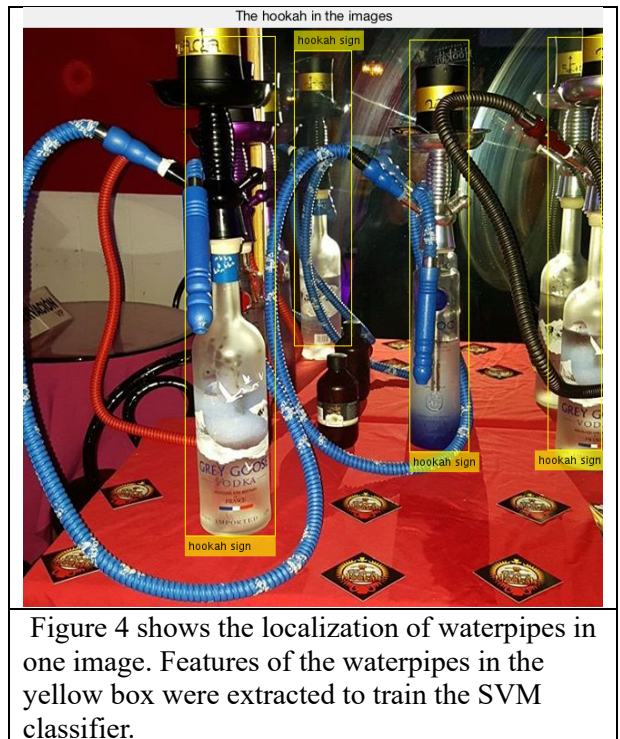

Figure 4 shows the localization of waterpipes in one image. Features of the waterpipes in the yellow box were extracted to train the SVM classifier.

## 3 Results:

Results demonstrated that hookah features could be extracted by CNN, with image categories classified by the SVM, maintaining a high level of accuracy (highest 99.5%). Figure 5 shows the features that were extracted from the first convolutional layer; this layer can only detect the edges and blobs, while more features were extracted from the remaining convolutional layers. Figure 6 shows the feature vectors of the 420 training images, with range [-20 20]; the majority of feature vectors are located between -10 and 10. Figure 7 is the histogram of the feature vectors. The maximum number of features was between -2 and 2. This interval [-2 2] reflected the most important features of the hookah images. Figure 8 shows the relationship between the function evaluations and the minimum objective. Function evaluations demonstrates how many times to evaluate the optimized output. The minimum objective is the minimum observed value of the objective function. It is the smallest overall observation point if there are coupled constraints or evaluation errors. The estimation of minimum objective functions, which can show the difference between the estimated (optimized) minimum objective and real minimum objective. The minimum objective and the estimated minimum objective are similar, however there are differences across certain function evaluations. The maximum proportion of error is less than 0.01, which is acceptable.[19] Based on the optimized SVM classifier, we evaluated the performance of our method by the test images.

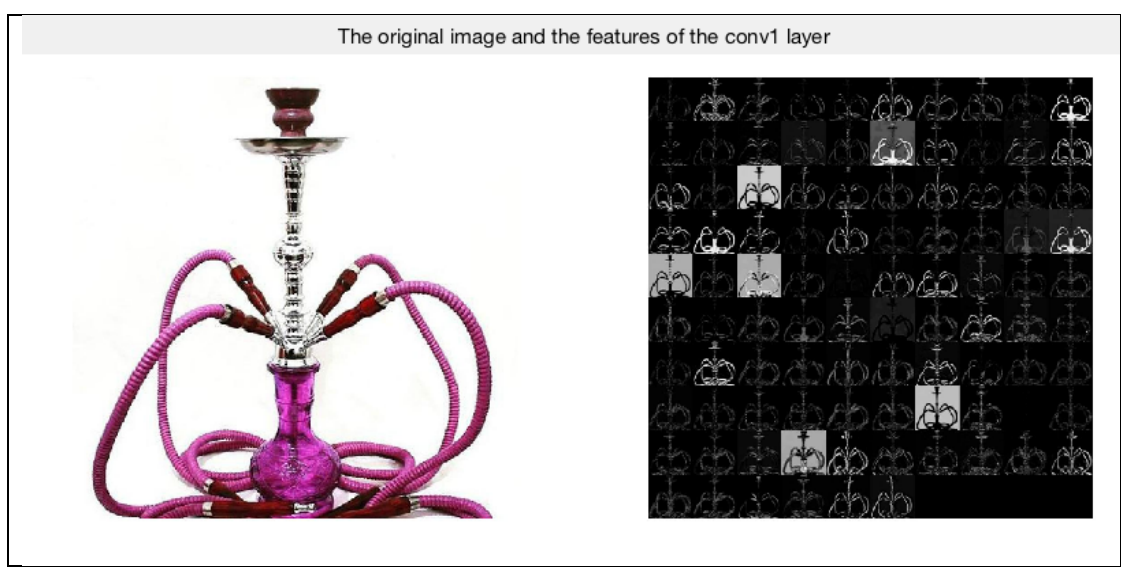

The original image and the features of the conv1 layer

Figure 5 shows the extracted features of the first layer using the CNN. The original hookah image is on the left. The feature images (right) contains a montage of 96 images, which can reflect the processing of extracting features.

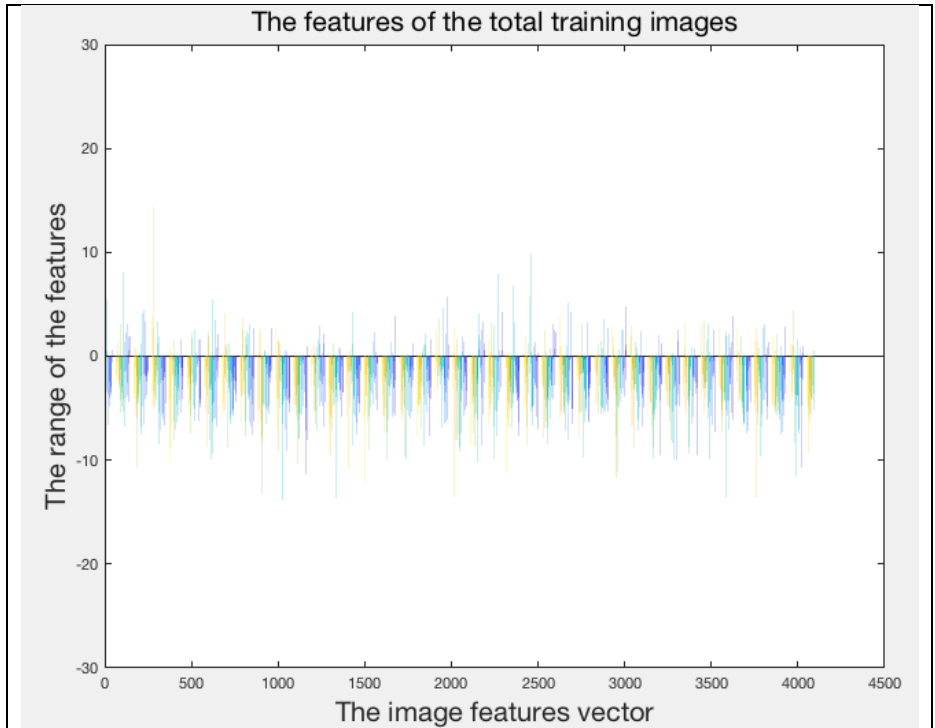

Figure 6 shows the features of the total image sets (420 images). The x-axis is the image features vector with 4096 total feature vectors. The y-axis is the range of the features with the range between -20 and 20.

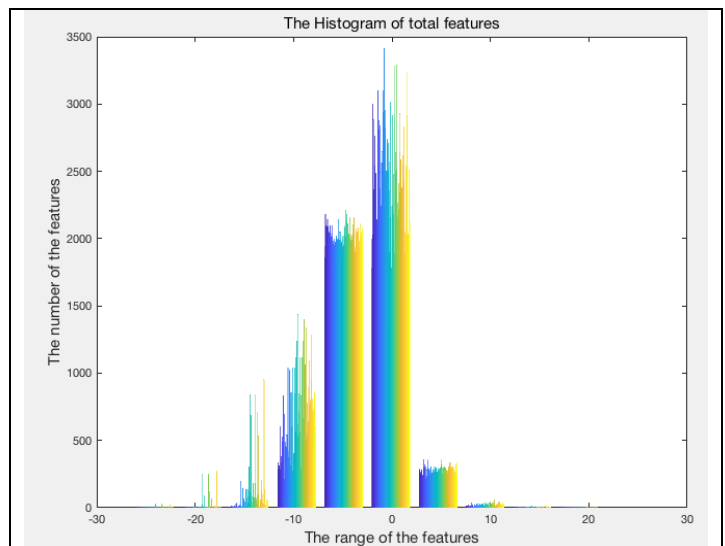

Figure 7 shows the histogram of the features. The interval of [-2 2] contains the maximum number of features.

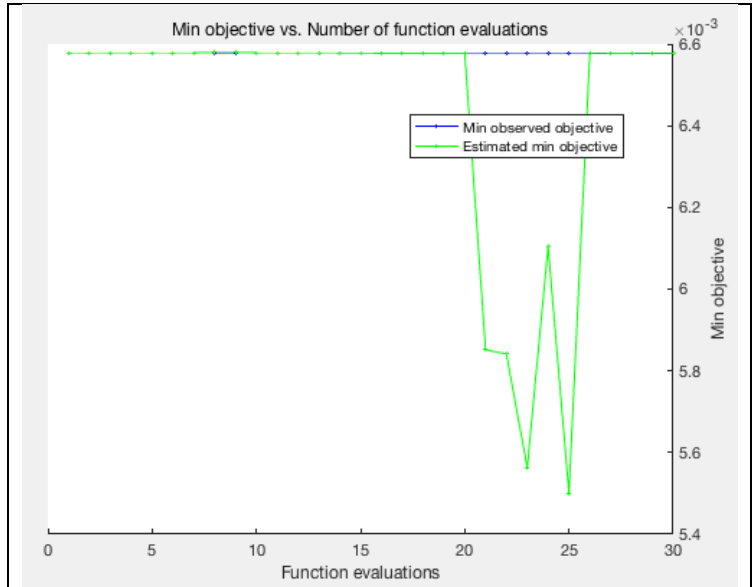

Figure 8 shows the relationship between the function evaluations and the minimum objective. When the function evaluation was 25, the error between minimum objective, and the estimated minimum objective was highest.

### 3.1 The test image classification results

Figure 9 is the learning curve which shows the relationship between the percentage of validated images (e.g., the training images, excluding the test images) and the average level of accuracy of the method. As the number of the validated training images increased, the total number of extracted features increased (For one image, we can extract 4096 features, therefore, with the number of the validated training images (n) increased, the total number of extracted features can increase into n×4096). Additionally, as the number of features learned by the SVM classifier increased, the average level of accuracy increased. The number of validated images was equal to the percentage × the number of the training set, e.g., if 10%, then the validated images=10% × 420=42.

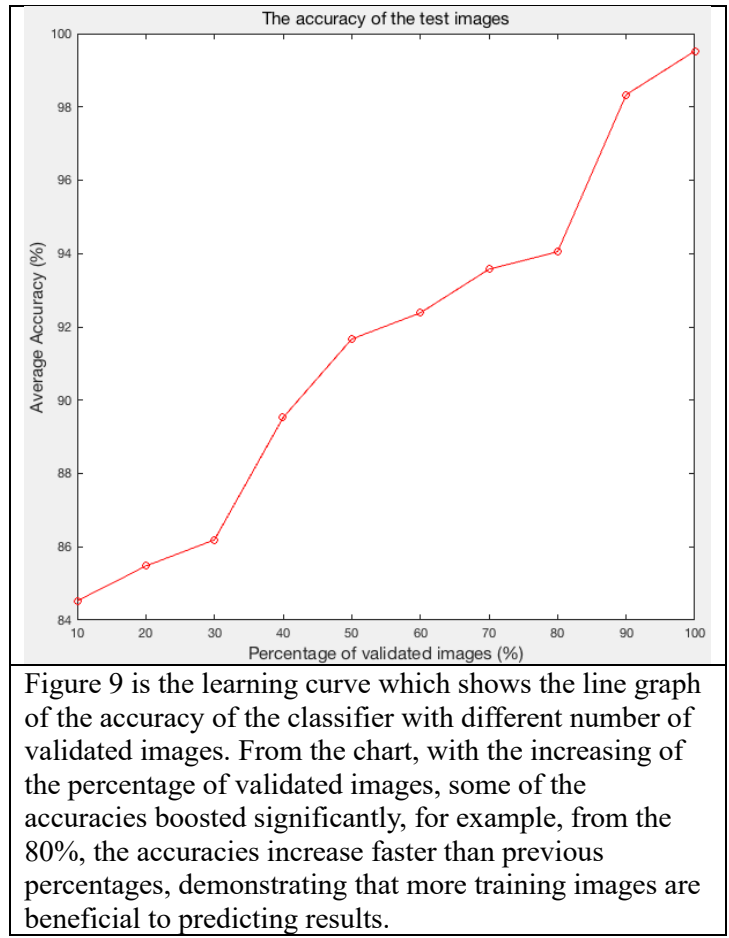

Figure 9 is the learning curve which shows the line graph of the accuracy of the classifier with different number of validated images. From the chart, with the increasing of the percentage of validated images, some of the accuracies boosted significantly, for example, from the 80%, the accuracies increase faster than previous percentages, demonstrating that more training images are beneficial to predicting results.

Overall, 99.5% of the 420 images classified were correctly identified as either hookah or non- hookah images (Figure 10). In the first row, all non-hookah images were correctly classified as such. In the second row, there were two hookah images incorrectly classified as non-hookah images, representing 3.4% of all the data. In the first row, 100% of hookah images were correctly classified. In the second row, 99.1%were correctly classified as hookah images. In the first column, 99% were correctly classified as hookah images and 0.1% were correctly classified as non-hookah images. In the second column, out of 210 non-hookah images, 100% were correctly classified as non-hookah images and all images were correctly classified as hookah images.

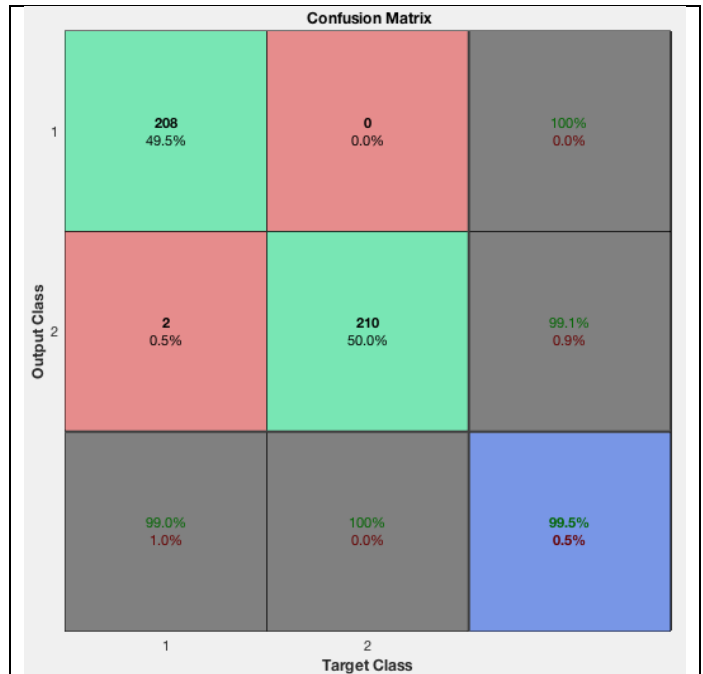

Figure 10 shows the confusion matrix of the test images (column 1 and 2 are the hookah and non-hookah categories, respectively, column 3 is the accuracy of classified results). The first two green squares show the number of the test images and the percentage of the correct image classifications. For example, there were 208 images correctly classified as hookah, and this number accounted for 49.5% of all test images (420). Similarly, 210 images were properly classified as non-hookah, and this accounted for 50% of all test images.

## 3.2 Compared with other methods

We compared our method with CNN, SVM, and BOF (see [35][36]for more details). For SVM and BOF, the input is the original image (raw pixel values). Figure 11 shows how the accuracy of various models can be improved as a function of the size of the training data. Our method had the highest accuracy (99.5%), compared to the other models.

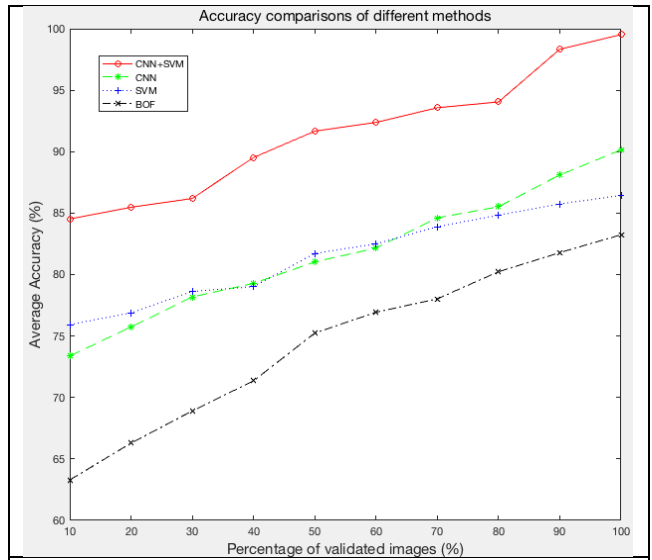

Figure 11. The predicting accuracy of different methods with different percentages of validated images (see definition in 3.1). CNN+SVM has higher accuracy than CNN, SVM and BOF.

## 4 Discussion

This study showed that the use of CNN to extract features and SVM to classify images results in higher accuracy in automated image classification compared to CNN or SVM alone. One crucial advantage of our pipelined approach is that we extracted enough features (4096 features from each image representing the details of each image) from a pretrained CNN model (AlexNet), taking advantage of SVM to train the features, saving time. Compared to earlier work using CNN, SVM, and BOF, our method improves accuracy when the number of training images is increased with accuracy approaching 100% (99.5%). This illustrates that our method is suitable for distinct images like waterpipes.

The methods presented here may be able to help detect increases in popularity of certain tobacco products over time on social media. By identifying waterpipes in images from Instagram we can identify Instagram users who may need tobacco-related education to curb hookah use. Instagram may be used to bolster the reach and delivery of health information that communicates the risk of hookah use[37][38][39][40]. Earlier research used Instagram images to capture and describe the context in which individuals use, and are marketed tobacco products Error! Reference source not found.[24].For example, analysis of Instagram data on electronic cigarettes demonstrated that a majority of images were either individuals showing their favorite combinations of products (e.g., type of electronic cigarette device and flavored juice), or people performing tricks with the products (e.g., blowing a large aerosol cloud in competition with others)[25], demonstrating how and why people use this tobacco product. Previous analyses of hookah-related posts to social media websites provide information about hookah-related contexts, including the importance of stylized waterpipes, use of hookah in social settings, co-promotion with alcohol[24], and primarily positive user experiences[41][42][43].

Earlier studies using image-based data provided timely information from a novel data source; however their methods relied upon hand coding of images--a process requiring time, expertise and sample sizes small enough to reasonably code by hand, ultimately limiting the scope of the work. The findings from the current study showed how automated image classification can be used to overcome such limitations. Additionally, the methods from the current study can help researchers in tobacco control identify what proportion of viewers on a social media site are interested in certain products such methods may be crucial to document the every changing tobacco landscape.

The findings from this study should be considered with several limitations in mind, including the fact that our task was a simple binary classification (hookah vs. non-hookah) which may result in high accuracy. To eliminate the problem of overfitting, we used ReLU, softmax, dropout layers in the CNN, and utilized several different training datasets (the number of datasets is different which increased from 42 to 420, in figure 11). The methods developed in this study were only applied in the context of images from Instagram that focused on waterpipes and should be applied in more categories and other contexts in the future. While we had high accuracy in classification, accuracy could be improved with better input features from the CNN model. In the future, researchers should try to enlarge the sets of training images to extract specific features of an image, which may achieve higher accuracy with less computation power.

**5 Conclusion**

Findings demonstrated that by combining CNN and SVM to classify images resulted in 99.5% accuracy in image classification, which is an improvement over earlier method using SVM, CNN or BOF alone. CNN extracts more features of the images allowing the SVM classifier to be better informed which results in higher accuracy compared with methods that extract fewer features. Future research can use our method to reduce computational time in identifying objects in images.

**Acknowledgments**

Research reported in this publication was supported by Grant # P50CA180905 from the National Cancer Institute and the FDA Center for Tobacco Products (CTP). The NIH or FDA had no role in study design, collection, analysis, and interpretation of data, writing the report, and the decision to submit the report for publication. The content is solely the responsibility of the authors and does not necessarily represent the official views of the NIH or FDA.

**Competing Interests**

None declared

**Author Contributions**

YZ and JPA conceived of the study and analyzed the data. YZ and JPA drafted the initial manuscript. JBU and TBC received funding for the study. JBU and TBC revised the manuscript for important intellectual content. All authors have approved the final manuscript.

**Reference:**

[1]     USA, San Francisco, CA: Instagram Business Team. (2016, SEPTEMBER 22). Hitting 500,000 Advertisers [Web log post]. Retrieved July 25, 2018, from https://business.instagram.com/blog/500000-advertisers/.

[2]    Ayers, J. W., Althouse, B. M., Leas, E. C., Dredze, M., & Allem, J. P. Internet searches for suicide following the release of'13 Reasons Why'. *Health*, 2003: *57*(4): 238-240.

[3]    Allem J, Escobedo P, Chu K, Boley Cruz T, Unger J. Images of Little Cigars and Cigarillos on Instagram Identified by the Hashtag #swisher: Thematic Analysis. *Journal of Medical Internet Research*. 2017;19(7):e255. PMID: 28710057.

[4]    Leas E, Althouse B, Dredze M, Obradovich N, Fowler J, Noar S et al. Big Data Sensors of Organic Advocacy: The Case of Leonardo DiCaprio and Climate Change. *PLOS ONE*. 2016;11(8): e0159885.

[5]    Perronnin, F., Sánchez, J., & Mensink, T. Improving the fisher kernel for large-scale image classification. *Computer Vision–ECCV*. 2010: 143-156.

[6]    Verma, K., Verma, L. K., & Tripathi, P. Image Classification using Backpropagation Algorithm. *Journal of Computer Science*, (2014); *1*(2).

[7]   Simonyan, K., Vedaldi, A., & Zisserman, A. Deep inside convolutional networks: Visualising image classification models and saliency maps. (2013). *arXiv preprint arXiv:1312.6034*.

[8]   Orlov N, Shamir L, Macura T, Johnston J, Eckley D, Goldberg I. WND-CHARM: Multi-purpose image classification using compound image transforms. *Pattern Recognition Letters*. 2008;29(11):1684-1693.

[9]   Karimy Dehkordy, Hossein. Automated image classification via unsupervised feature learning by K-means. *Diss*. 2015.

[10]  Sun J, Yang J, Zhang C, Yun W, Qu J. Automatic remotely sensed image classification in a grid environment based on the maximum likelihood method. *Mathematical and Computer Modelling*. 2013; 58(3-4): 573-581.

[11]  Chapelle, O., Haffner, P., & Vapnik, V. N. Support vector machines for histogram-based image classification. *IEEE transactions on Neural Networks*. 1999; *10*(5): 1055-1064.

[12]  Nowak, E., Jurie, F., & Triggs, B. Sampling strategies for bag-of-features image classification. *Computer Vision–ECCV*. 2006; 490-503.

[13]  Cireşan D, Meier U, Masci J, Schmidhuber J. Multi-column deep neural network for traffic sign classification. *Neural Networks*. 2012; 32:333-338.

[14]  Jiang, Y. G., Ngo, C. W., & Yang, J. Towards optimal bag-of-features for object categorization and semantic video retrieval. In *Proceedings of the 6th ACM international conference on Image and video retrieval*. ACM. 2007: 494-501.

[15]  Marszaek, M., & Schmid, C. Spatial weighting for bag-of-features. In Computer Vision and Pattern Recognition, *2006 IEEE Computer Society Conference on IEEE*. 2006; 2: 2118-2125.

[16]  Auria, L., & Moro, R. (2007). Advantages and Disadvantages of Support Vector Machines. Credit Risk Assessment Revisited: *Methodological Issues and Practical Implications*, 49-68.

[17]  Burges, C. J. A tutorial on support vector machines for pattern recognition. *Data mining and knowledge discovery*. 1998; *2*(2): 21-167.

[18]  Furey, T. S., Cristianini, N., Duffy, N., Bednarski, D. W., Schummer, M., & Haussler, D. Support vector machine classification and validation of cancer tissue samples using microarray expression data. *Bioinformatics*. 2000; *16*(10): 906-914.

[19]  Simonyan, K., & Zisserman, A. Very deep convolutional networks for large-scale image recognition. 2014. *arXiv preprint arXiv:1409.1556*.

[20]  Vedaldi, A., & Lenc, K. Matconvnet: Convolutional neural networks for matlab. In *Proceedings of the 23rd ACM international conference on Multimedia*, ACM. 2015: 689-692.

[21]  Ayers J, Leas E, Allem J, Benton A, Dredze M, Althouse B et al. Why do people use electronic nicotine delivery systems (electronic cigarettes)? A content analysis of Twitter, 2012-2015. *PLOS ONE*. 2017;12(3): e0170702.

[22]  Ayers, J. W., Leas, E. C., Dredze, M., Allem, J. P., Grabowski, J. G., & Hill, L. Pokémon GO—a new distraction for drivers and pedestrians. *JAMA internal medicine*. 2016; *176*(12): 1865-1866.

[23] Allem J, Leas E, Caputi T, Dredze M, Althouse B, Noar S et al. The Charlie Sheen Effect on Rapid In-home Human Immunodeficiency Virus Test Sales. *Prevention Science*. 2017;18(5):541-544.

[24] Allem J, Chu K, Cruz T, Unger J. Waterpipe Promotion and Use on Instagram: #Hookah. *Nicotine & Tobacco Research*. 2017;19(10):1248-1252. PMID: 28077449.

[25] Chu K, Allem J, Cruz T, Unger J. Vaping on Instagram: cloud chasing, hand checks and product placement. *Tobacco Control*. 2016;26(5):575-578.

[26] V. Nair and G. E. Hinton. Rectified linear units improve restricted boltzmann machines. In Proc. 27th, *International Conference on Machine Learning* (ICML-10). 2010: 807-814.

[27] Ballester, P., & de Araújo, R. M. (2016, February). On the Performance of GoogLeNet and AlexNet Applied to Sketches. In *AAAI* (pp. 1124-1128).

[28] Yu, W., Yang, K., Bai, Y., Xiao, T., Yao, H., & Rui, Y. (2016, June). Visualizing and comparing AlexNet and VGG using deconvolutional layers. *In Proceedings of the 33 rd International Conference on Machine Learning*.

[29] Krizhevsky, Alex, Ilya Sutskever, and Geoffrey E. Hinton. "Imagenet classification with deep convolutional neural networks." *Advances in neural information processing systems*. 2012: 1097-1105.

[30] G.E. Hinton, N. Srivastava, A. Krizhevsky, I. Sutskever, and R.R. Salakhutdinov. Improving neural net-works by preventing co-adaptation of feature detectors. 2012. *arXiv preprint arXiv*:1207.0580.

[31] Hua, S., & Sun, Z. Support vector machine approach for protein subcellular localization prediction. *Bioinformatics*. 2001; 17(8): 721-728.

[32] Furey, T. S., Cristianini, N., Duffy, N., Bednarski, D. W., Schummer, M., & Haussler, D. Support vector machine classification and validation of cancer tissue samples using microarray expression data. *Bioinformatics*. 2000; 16(10): 906-914.

[33] Escalera S, Pujol O, Radeva P. On the Decoding Process in Ternary Error-Correcting Output Codes. *IEEE Transactions on Pattern Analysis and Machine Intelligence*. 2010;32(1):120-134.

[34] Escalera S, Pujol O, Radeva P. Separability of ternary codes for sparse designs of error-correcting output codes. *Pattern Recognition Letters*. 2009;30(3):285-297.

[35] Nowak, E., Jurie, F., & Triggs, B. Sampling strategies for bag-of-features image classification. *Computer Vision–ECCV*. 2006: 490-503.

[36] égou, H., Douze, M., & Schmid, C. Improving bag-of-features for large scale image search. *International journal of computer vision*. 2010; 87(3): 316-336.

[37] Pechmann C, Pan L, Delucchi K, Lakon C, Prochaska J. Development of a Twitter-Based Intervention for Smoking Cessation that Encourages High-Quality Social Media Interactions via Automessages. *Journal of Medical Internet Research*. 2015;17(2): e50. PMID: 25707037.

[38] Pechmann C, Delucchi K, Lakon C, Prochaska J. Randomised controlled trial evaluation of Tweet2Quit: a social network quit-smoking intervention. *Tobacco Control*. 2016;26(2):188-194. PMID: 26928205.

[39] Allem J, Escobedo P, Chu K, Soto D, Cruz T, Unger J. Campaigns and counter campaigns: reactions on Twitter to e-cigarette education. *Tobacco Control*. 2016;26(2):226-229. PMCID: PMC5018457.

[40] Naslund JA, Kim SJ, Aschbrenner KA, McCulloch LJ, Brunette MF, Dallery J, Bartels SJ, Marsch LA. Systematic review of social media interventions for smoking cessation. *Addict Behav*. 2017; 73: 81-93. PMID: 28499259.

[41] Chen, A. T., Zhu, S., & Conway, M. (2015). Combining Text Mining and Data Visualization Techniques to UnderstandConsumer Experiences of Electronic Cigarettes and Hookah in OnlineForums. *Online journal of public health informatics*, 7(1).

[42] Krauss, M. J., Sowles, S. J., Moreno, M., Zewdie, K., Grucza, R. A., Bierut, L. J., & Cavazos-Rehg, P. A. (2015). Peer reviewed: Hookah-related twitter chatter: A content analysis. *Preventing chronic disease*, 12.

[43] Myslín M, Zhu SH, Chapman W, Conway M. Using twitter to examine smoking behavior and perceptions of emerging tobacco products. *J Med Internet Res*. 2013;15:e174. PMID: 23989137.